\documentclass{article}

\PassOptionsToPackage{numbers, compress}{natbib}

    \usepackage[preprint]{neurips_2021_ml4ad}

\usepackage{natbib}
\usepackage{floatrow} 
\usepackage{graphics} 
\usepackage{epsfig} 
\usepackage[tight]{subfigure}
\usepackage{bm}
\usepackage{tabularx,pbox,booktabs,caption}
\usepackage{multirow}
\usepackage{svg}
\usepackage[utf8]{inputenc} 
\usepackage[T1]{fontenc}    
\usepackage{url}            
\usepackage{booktabs}       
\usepackage{amsfonts}       
\usepackage{nicefrac}       
\usepackage{microtype}      
\usepackage{xcolor}         

\title{NSS-VAEs: Generative Scene Decomposition for Visual Navigable Space Construction}

\author{%
  Zheng Chen, Lantao Liu
\\
  Luddy School of Informatics, Computing, and Engineering\\
  Indiana University Bloomington\\
  \texttt{\{zc11, lantao\}@iu.edu} \\
}

\begin{document}

\maketitle

\vspace{-20pt}
\begin{abstract}
Detecting navigable space is the first and also a critical step for successful robot navigation. In this work, we treat the visual navigable space segmentation as a scene decomposition problem and propose a new network, NSS-VAEs (\textbf{N}avigable \textbf{S}pace \textbf{S}egmentation \textbf{V}ariational \textbf{A}uto\textbf{E}ncoders), a representation-learning-based framework to enable robots to learn the navigable space segmentation in an unsupervised manner. Different from prevalent segmentation techniques which  heavily rely on supervised learning strategies and typically demand immense pixel-level annotated images, the proposed framework leverages a generative model -- Variational Auto-Encoder~(VAE) -- to learn a probabilistic polyline representation that compactly outlines the desired navigable space boundary.
Uniquely, our method also assesses the prediction uncertainty related to the unstructuredness of the scenes, which is important for robot navigation in unstructured environments. Through extensive experiments, we have validated that our proposed method can achieve remarkably high accuracy ($>90\%$) even without a single label. We also show that the prediction of NSS-VAEs can be further improved using few labels with results significantly outperforming the SOTA fully supervised learning-based method.
\vspace{-5pt}
\end{abstract}

\vspace{-5pt}
\section{Introduction and Related Work} \vspace{-5pt}
\label{sect:intro}
\vspace{-5pt}
A crucial capability for mobile robots to navigate in unknown environments is to construct obstacle-free space where the robot could move without collision. Roboticists have been developing methods for detecting such free space with the ray trace of LiDAR beams to build occupancy maps in 2D or 3D space~\cite{thrun2003learning, senanayake2017bayesian}.  Mapping methods with LiDAR require processing of large point cloud data, especially when a high-resolution LiDAR is used. 
As a much less expensive alternative, cameras have also been tremendously used for free space detection by leveraging deep neural networks (DNNs), to perform multi-class~\cite{tian2019decoders, wang2018depth, yang2018denseaspp, wigness2019rugd} or binary-class~\cite{tsutsui2018minimizing, fan2020sne, tsutsui2017distantly, pizzati2019enhanced, wang2020applying, kothandaraman2020safe} segmentation of images. 

However, most of existing DNN based methods are built on a supervised-learning paradigm and rely on annotated datasets (e.g., KITTI~\cite{geiger2013vision} and CityScape~\cite{cordts2015cityscapes}). The datasets usually contain a large amount of pixel-level annotated segmented images, which are prohibitively expensive and time-consuming to obtain for many robotic applications. To overcome limitations of fully-supervised learning, we develop a self-supervised learning method by treating the binary navigable space segmentation  (i.e., navigable vs. non-navigable) as a scene decomposition problem. 

Specifically, we propose a model, NSS-VAEs (Navigable Space Segmentation VAEs) to perform visual navigable space segmentation in an unsupervised way. The model architecture consists of two VAEs, VAE-I and VAE-II, where the goal of VAE-I is to learn a pseudo label from surface normal images by using categorical distributions as latent representations. Using supervision signals from VAE-I, VAE-II then can learn a set of Gaussian distributions which describe the position as well as the structural uncertainty of the navigable space boundary (for assessment of environmental unstructureness). The proposed framework pipeline is shown in Figure.~\ref{fig:system}.

\vspace{-5pt}
\section{Navigable Space Segmentation with Two Layers of VAEs} \vspace{-5pt}
\label{sect:method}

\begin{figure} \vspace{-15pt}
\centering
  {\includegraphics[width=0.7\linewidth]{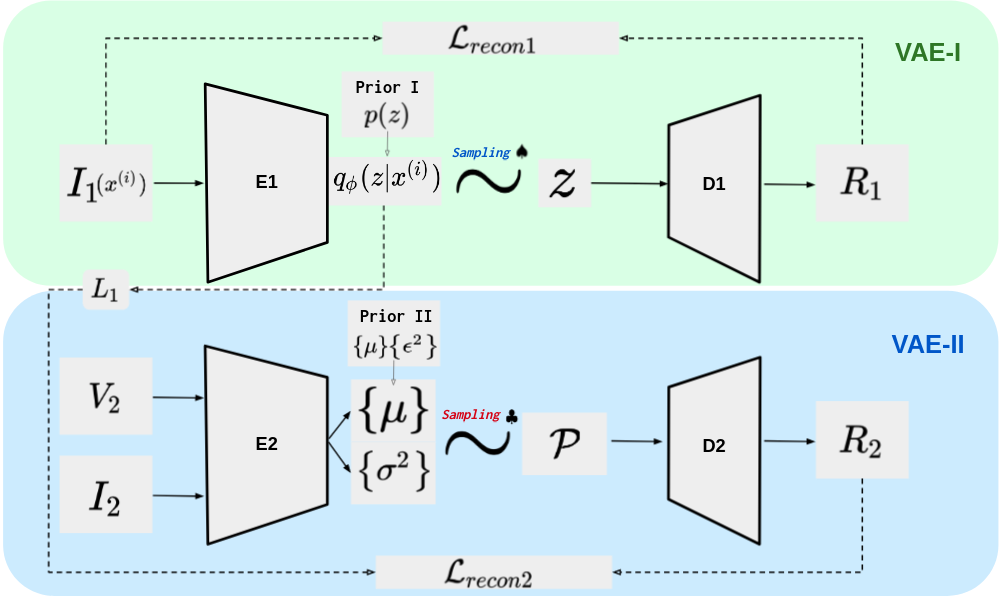}}
\caption{\small Framework overview. \textbf{VAE-I}: $I_1$ is the input surface normal image. $q_{\phi}(z|x^{(i)})$ represents the latent categorical distribution, from which a segmentation sample $z$ is sampled using $\spadesuit$, the Gumbel-Softmax sampler for discrete distributions proposed in \cite{jang2016categorical}. Prior I is to regularize the shape of the predicted latent distribution. $R_1$ is the reconstructed input and $\mathcal{L}_{recon1}$ is the mean squared error between $I_1$ and $R_1$. \textbf{VAE-II}: $I_2$ is the same as $I_1$. $V_2$ is the initial vertices for polyline representation. $\left \{ \mu \right \}$ and $\left \{ \sigma^2 \right \}$ are sets of parameters of Gaussians, representing distributions of vertices of the predicted polyline. $\mathcal{P}$ is a set of sampled vertices using $\clubsuit$, the sampling operation for continuous distributions proposed in \cite{kingma2013auto}. Prior II is to regularize the value of the predicted variances. $R_2$ is the reconstructed image and $\mathcal{L}_{recon2}$ is the mean squared error between $R_2$ and the latent image $L_1$, converted from the latent distribution $q_{\phi}$.
\vspace{-10pt}
} 
\label{fig:system}  
\end{figure}

A standard VAE aims to learn a generative model which could generate new data from a random sample in a specified latent space. The model parameters are optimized by maximizing the data marginal likelihood: $\log p_{\theta}(\mathbf{x}^{(i)})$, where $\mathbf{x}^{(i)}$ is one of data points in our training dataset $\left \{ \mathbf{x}^{(i)} \right \}_{i=1}^N$. Using Bayes' rule, the likelihood could be written as:
\begin{equation}
\label{eq:marginal_likelihood}
    \log p_{\theta}(\mathbf{x}^{(i)}) = \log \sum_{\mathbf{z}} p_{\theta}(\mathbf{x}^{(i)}|\mathbf{z})p_{\theta}(\mathbf{z}),
\end{equation}
where $p_{\theta}(\mathbf{z})$ is the prior distribution of the latent representation; $p_{\theta}(\mathbf{x}^{(i)}|\mathbf{z})$ is the generative probability distribution of the
reconstructed input given the latent representation. \\
\textbf{VAE-I:} To perform the segmentation task, we approximate the posterior distribution  
$q_{\phi}(\mathbf{z}|\mathbf{x}^{(i)}) = \prod_{j=1}^J Cat(z_j|\mathbf{x}^{(i)}, \phi)$, and assume the prior $p_{\theta}(\mathbf{z}) = \prod_{j=1}^J Cat(z_j)$, where $J$ is the number of pixels in the input image (same meaning in following sections). 
Then the \textit{Evidence Lower BOund}~(ELBO) of the data marginal likelihood can be derived as:
\begin{equation}
    \label{eq:elbo}
    ELBO = \mathbb{E}_{\mathbf{z}} \log p_{\theta} (\mathbf{x}^{(i)}|\mathbf{z}) - \mathbb{KL}(q_{\phi}(\mathbf{z}|\mathbf{x}^{(i)}) || p_{\theta}(\mathbf{z})),
\end{equation}
where $p_{\theta}(\mathbf{x}^{(i)}|\mathbf{z})$ is the generative probability distribution of the reconstructed input given the latent representation and we assume $p_{\theta}(\mathbf{x}^{(i)} | \mathbf{z}) = \prod_{j=1}^J \mathcal{N} (\mathbf{x}^{(i)}; \hat{\mathbf{x}}_j(\mathbf{z};\theta), \sigma^2)$. By utilizing Monte Carlo sampling, the total loss function over the whole training dataset is:
\begin{equation}
    \label{eq:loss_1}
    \mathcal{L}_1 = \sum\nolimits_{i=1}^N  \mathbb{KL}(q_{\phi}(\mathbf{z}|\mathbf{x}^{(i)}) || p_{\theta}(\mathbf{z})) + \frac{1}{2K\sigma^2}\sum\nolimits_{k=1}^K \left \| \mathbf{x}^{(i)} - \hat{\mathbf{x}}(\mathbf{z}_{ik}; \theta) \right \|_2^2  + \frac{J}{2} \log \sigma^2,
\end{equation}
where $N$ is the number of images in our training dataset and $K$ is the number of Monte Carlo samples for each image.\\
\begin{figure} \vspace{-15pt}
\centering
  {\includegraphics[width=0.7\linewidth]{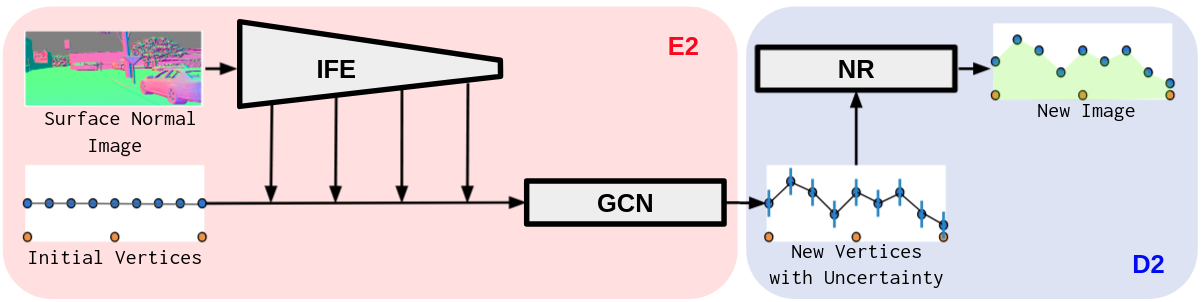}}
\caption{\small Structure of VAE-II. \textit{Left block}: encoder of VAE-II. IFE: Image Feature Extraction net. GCN: Graph convolutional network. \textit{Right block}: decoder of VAE-II. NR: Neural rendering module. In both blocks, blue points are vertices of polyline while the orange points are auxiliary points on image bottom boundary for the convenience of neural rendering.} 
\vspace{-10pt}
\label{fig:vae2}  
\end{figure}
\begin{figure} \vspace{-11pt}
\centering
  {\includegraphics[width=0.7\linewidth]{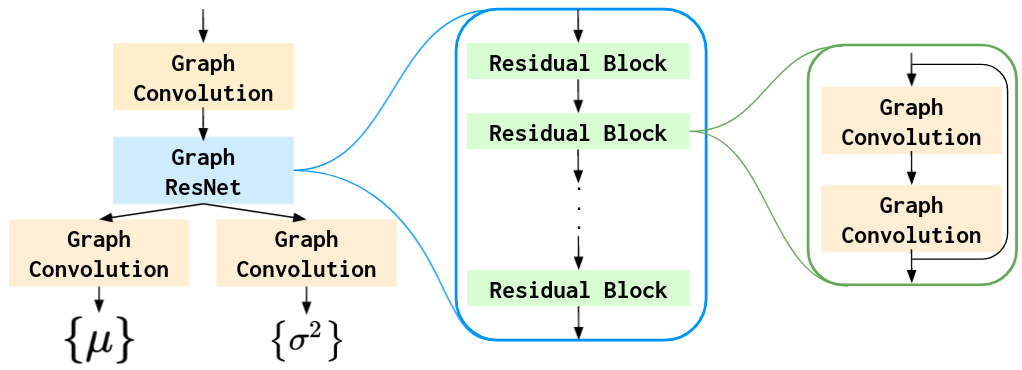}} \vspace{-8pt}
\caption{\small GCN module in the \textit{left block} of Figure.~\ref{fig:vae2}. 
}  \vspace{-4pt}
\label{fig:gcn}  
\end{figure}
\textbf{VAE-II: }
A more specific pipeline of VAE-II can be seen in Figure.~\ref{fig:vae2}. The IFE module is used to extract deep features of different layers from the input image for each vertex. We construct a graph using the vertices and use the concatenation of the extracted image features and the coordinates as the feature of each node in the graph. Our GCN structure (see Figure.~\ref{fig:gcn}) is inspired by the network structure proposed in recent work \cite{wang2018pixel2mesh, ling2019fast}. 

To improve the model representation capability, we propose to regularize the predicted variances with prior distributions. The distance between the two sets of distributions are:
\begin{equation}
    \label{eq:kl_prior}
    D_{KL}(\mathcal{N}_1, \mathcal{N}_2) = \frac{1}{N}\sum_{i=1}^N \log \epsilon_i - \log \sigma_i + \frac{\sigma_i^2 - \epsilon_i^2}{2\epsilon_i^2},
\end{equation}
where $\mathcal{N}_1(\bm{\mu}, \bm{\sigma}^2)$ is the predicted distributions; $\mathcal{N}_2(\bm{\mu}, \bm{\epsilon}^2)$ is the specified prior distributions; and $N$ is the number of vertices. We achieve the regularization by adding Eq.~(\ref{eq:kl_prior}) to the loss of VAE-II.

Sampling from the predicted distributions can give us a set of vertices $\mathcal{P}$ (see Figure.~\ref{fig:system}). To convert the vertices to a reconstructed image, we then triangularize those vertices and select proper triangles for neural rendering.\\
\begin{figure} \vspace{-10pt}
\centering
  {\includegraphics[width=\linewidth]{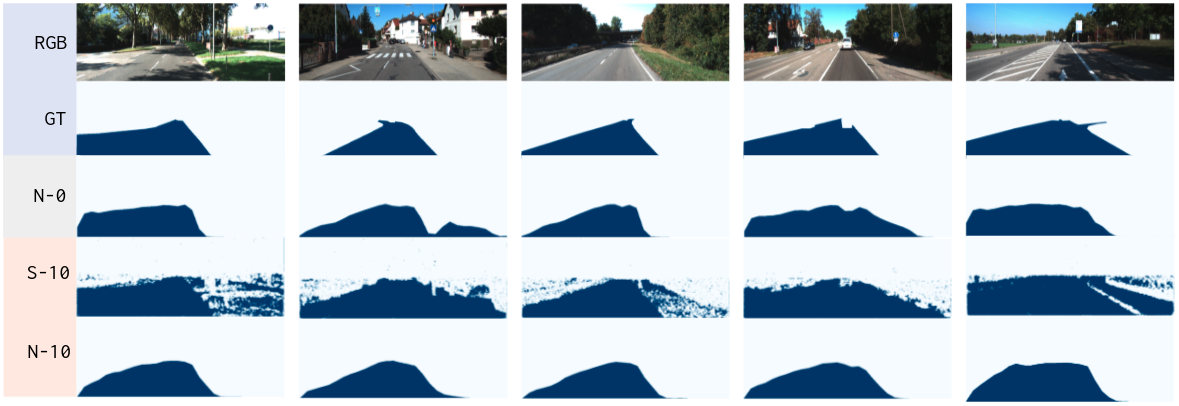}}
\caption{\small Qualitative results on KITTI road benchmark. We show examples of the predicted segmentation on testing data. First row: RGB images; Second row: ground-truth navigable space segmentation. Label for each row (from the third to the last one), S/N-$\#$: S represents the baseline method, SNE-RoadSeg~\cite{fan2020sne}; N represents the proposed method, NSS-VAEs; $\#$ represents the percentage of the used gt labels. Note: the baseline method will not work if no gt label is provided, thus S-0 is unavailable.} 
\vspace{-10pt}
\label{fig:images}  
\end{figure}
\textbf{Self-Supervised Objective: }
Similar to Eq.~(\ref{eq:loss_1}), the overall objective of VAE-II consists of a reconstruction loss and a prior loss $\mathcal{L}_2 = \mathcal{L}_{recon2} + \mathcal{L}_{prior}$, 
where 
$\mathcal{L}_{prior}$ is the KL divergence between two distributions (Eq.~\ref{eq:kl_prior}). Inspired by the appearance matching loss in \cite{guizilini20203d}, $\mathcal{L}_{recon2}$ is estimated using the Structural Similarity Index Measure (SSIM) combined with a Mean Squared Error (MSE) between $L_1$ and $R_2$.
We can write
$\mathcal{L}_2$ as:
\begin{equation}
    \label{loss_2_new}
    \mathcal{L}_2 = \lambda_1 \frac{1 - SSIM(L_1, R_2)}{2} + \lambda_2 \frac{1}{J}\left \|  L_1 - R_2\right \|^2_2 + \lambda_3 D_{KL}(\mathcal{N}_1, \mathcal{N}_2),
\end{equation}
where $J$ is the total number of pixels, $\lambda_1$, $\lambda_2$ and $\lambda_3$ are weights and $\lambda_1 + \lambda_2 + \lambda_3 = 1$.

\vspace{-5pt}
\section{Evaluations} 
\label{sect:exp}
\vspace{-12pt}
\begin{figure}
  \centering
    \subfigure
  	{\label{fig:accuracy}\includegraphics[width=0.18\linewidth]{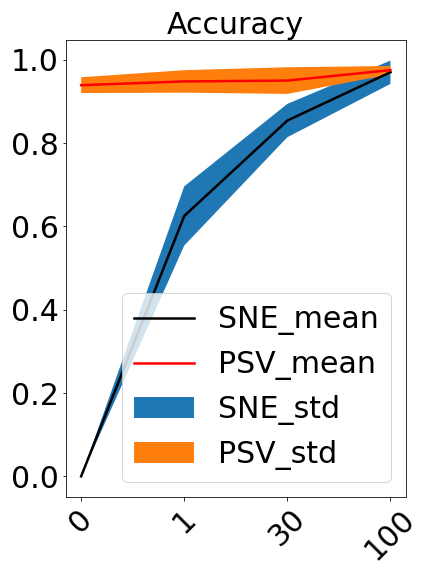}}
   	\subfigure
   	{\label{fig:precision}\includegraphics[width=0.18\linewidth]{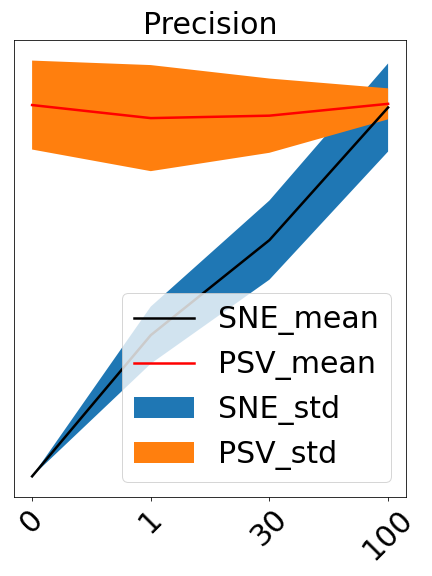}} 
   	\subfigure
   	{\label{fig:recall}\includegraphics[width=0.18\linewidth]{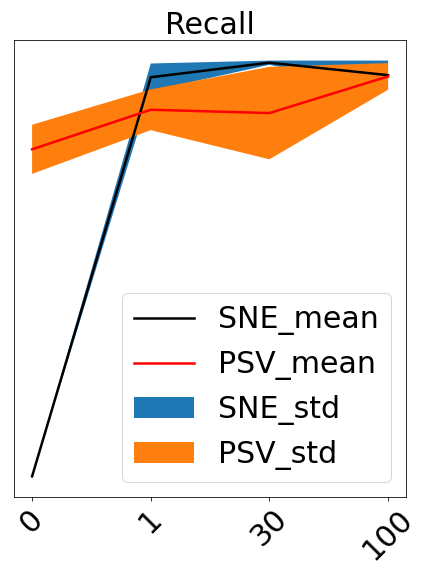}}
   	\subfigure
   	{\label{fig:f_score}\includegraphics[width=0.18\linewidth]{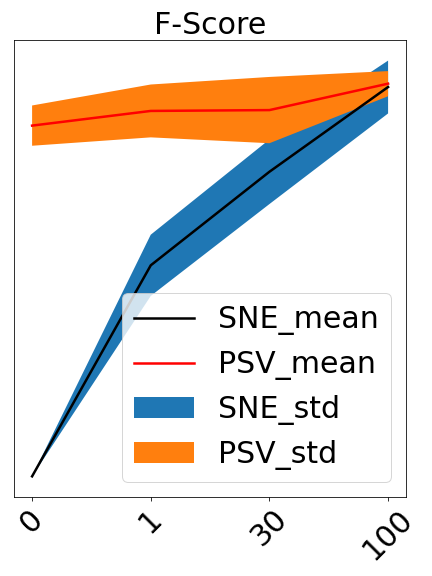}}
   	\subfigure
   	{\label{fig:iou}\includegraphics[width=0.18\linewidth]{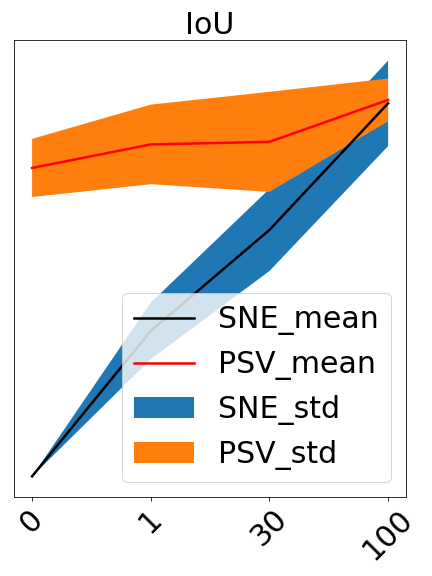}  }\vspace{-10pt}
   \caption{\small Quantitative results on KITTI road benchmark. The horizontal axis represents the percentage of the used gt labels (only the data of 4 percentages are shown: 0, 1, 30, and 100). The statistical results are computed from all of the 40 images in testing data. \vspace{-10pt}
   }
\label{fig:curves}  
\end{figure}

\begin{figure}\vspace{10pt}
\centering
  {\includegraphics[width=0.9\linewidth]{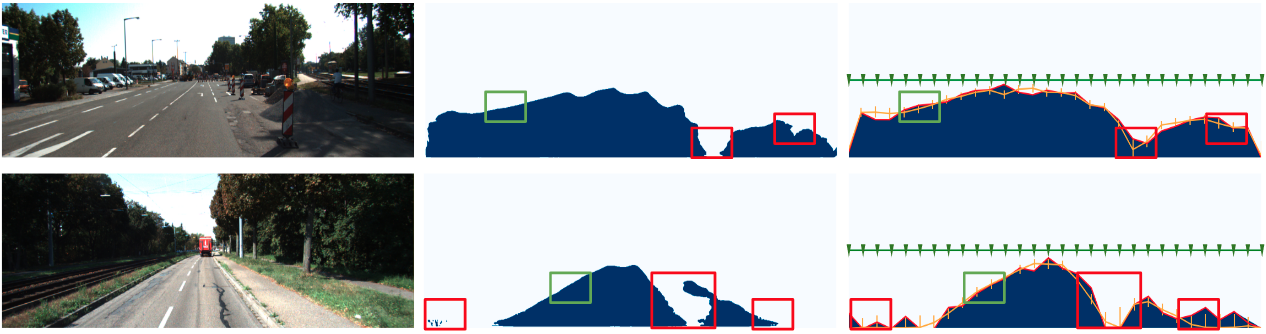}}
\caption{\small \textit{Left column}: RGB images; \textit{Middle column}: Noisy latent representation of VAE-I. Green blocks are image regions with few structural noises while red blocks are places with more structural noises; \textit{Right column}: Navigable space prediction from VAE-II. The green inverted triangles are the initial vertices. The orange line is the line connecting sampled vertices from the predicted distributions (mean and std are shown with error bar). 
} \vspace{-10pt}
\label{fig:uncertainty}  
\end{figure}

\textbf{Evaluation Metrics:}
\label{sect:eval_metrics}
We compare the most recent baseline method~\cite{fan2020sne} which is a fully supervised learning based SOTA approach for predicting navigable space. We use the same metrics adopted in \cite{fan2020sne}: \textbf{A}ccuracy, \textbf{P}recision, \textbf{R}ecall, \textbf{F}-Score, and \textbf{I}oU~(Intersection over Union). More details about the metrics could be seen in \cite{fan2020sne}.\\
\textbf{Segmentation Performance:}
A qualitative comparison between the proposed method with the baseline method is shown in Figure.~\ref{fig:images}, where predictions are shown with 2 different percentages of gt labels: $0\%$ and $10\%$. We can see the proposed NSS-VAEs can predict results close to the gt labels even without any gt labels when training (the third row). 
A more precise comparison can be seen in Figure.~\ref{fig:curves}. 
From the curves, it can be easily seen (from the mean) that the segmentation performance of both models are proportional to the number of gt labels. However, the accuracy (as well as other metrics) of NSS-VAEs can always stay at a high level (e.g., $>90\%$) while the baseline method tend to generate unacceptable results (e.g., $<80\%$) if not enough gt labels are available. 

Another advantage of our proposed NSS-VAEs is we can discretize the boundary of the navigable space into a series of vertices and treat the position of each vertex as a distribution.
By doing so, we are able to describe the structural noises due to ambiguities of visual information using the predicted uncertainty. The ambiguity may be due to several factors, e.g., light condition and textures. 
We show two examples of uncertainty prediction in Figure.~\ref{fig:uncertainty}. In the regions enclosed by red blocks, there are more noises due to (a) in the top image, e.g., the shadow, and the existence of the traffic cone and (b) in the bottom image, e.g., the shadow and the similar texture of the road and the sidewalk. The corresponding vertices (those enclosed by red blocks in the \textit{Right} images) have higher variances. In contrast, the noises are smaller in the regions enclosed by green blocks and so are the variances.  

\vspace{-15pt}
\section{Conclusion}
\vspace{-5pt}
\label{sect:conclusion}

We consider the challenging visual navigable space learning problem where pixel-wise annotated labels used by  fully supervised learning are expensive and time-consuming to obtain for robotics applications. 
To tackle this challenge, we propose a new network -- NSS-VAEs, to learn the navigable space in an unsupervised way. The proposed NSS-VAEs discretizes the boundary of navigable spaces into a set of vertices and leverages a generative model, VAEs, to learn probabilistic distributions of those vertices, thus our model is also able to assess the structural uncertainty of the scenes. With  extensive evaluations, we have validated the effectiveness of the proposed method and the remarkable advantages over the SOTA fully supervised learning baseline method.

\bibliography{ref}

\end{document}